# Disease Labeling via Machine Learning is NOT quite the same as Medical Diagnosis

Moshe BenBassat

Arison School of Business, Interdisciplinary Center (IDC), Herzliya, Israel

(www.moshebenbassat.com, email: moshe.benbassat@plataine.com)

## 1.      Disorder Labels in Medical Diagnosis

A key step in medical diagnosis is giving the patient's situation a *label*, e.g. Myocardial Infarction (MI, heart attack), or Appendicitis, which essentially assigns the patient to a class (or classes) of patients that are likely to have similar body failures and manifest similar findings (symptoms, history, physical examination, test results, etc.). Using the terminology of Artificial Intelligence (AI), this step is known as *labeling* and the medical findings are known as *features*. The classification label is not a free text written on the fly, but rather is taken from an approved list of hundreds of different disorder names, see ICD-10 2019, which have been compiled and refined by the medical profession over many years, e.g. K35.80 code for Appendicitis. In the early/intermediate stages of a diagnostic process, the list of candidate disorder labels that may account for the known set of findings is known by physicians as *differential diagnosis*, and the objective is to collect more findings, in a cost-effective manner, to converge to a situation where one or more labels are almost certain, while all others are virtually eliminated.

It should be stressed, however, that two patients may be assigned the same label, say D17, with roughly the same probability, say >85%, (and every expert would agree that D17 is the right one), but with important differences in their list of findings that may suggest *different ensuing actions*, such as deviations from the template treatment plan for D17. Feinstein's classic book on Clinical Judgement [1] contains excellent material on the foundations of medical diagnosis and disease categorization by one of the giants in medical thinking.

## 2.      Key Limitations of Disorder Labeling by Case-Based Machine Learning Algorithms

AI-based software for diagnosis support started 40+ years ago and is now re-emerging with great progress due to machine learning's evolution and breakthroughs, as well as increased data availability. Using AI has great potential to improve healthcare across the entire spectrum from clinical sessions to remote medicine. This article points at directions to improve the quality, depth and breadth of AI-based solutions for diagnosis support.  I was fortunate to be one of the pioneers in this space developing Bayesian Network algorithms for multiple co-existing disorders that were the AI core of MEDAS [2], [3]; a large-scale system for Emergency and Critical Care. In 1990, long after I moved on, MEDAS reached a "90% level agreement with the gold standard diagnosis of the attending physician" [4] and grew to cover **350 disorders** by means of **6000 features**.  More recent works on AI software are based on deep learning, random forest, gradient boosting, decision trees, or other **case-based** machine learning (ML) techniques. Evidence so far, from early and from recent works, strongly suggests that data-centric, and/or probability-centric, AI algorithms can reach expert level performance in the *disorder labeling phase*. However, these AI algorithms only make statistical class membership decisions; they **do not** 'understand' clinically what's going on with the patient in terms of physiology/anatomy to the level required to decide on the right ensuing actions. Findings are represented just by codes, say X(21) for Blood Pressure, X(72) for ST Elevation in EKG, and there are no software representations behind them of the body failures that produce these findings, such as representation of the human heart with its four chamber structure,... Pure case-based ML algorithms are **knowledge-blind**.

Statistically, a typical disorder is characterized by 30 to 90 most relevant features. For any given disorder, relatively few (<5%) of its characterizing features are absolutes ('Must be' or 'Cannot be'), and for most individual features there is more than one disorder that may account for them. This is reflected in their sensitivity/specificity values that are somewhere between zero and one (*sensitivity* and *specificity* are related terms to true positives, false positives). Assuming that for a given disorder, 60 of its characterizing features have sensitivity/specificity values in that range, it means that there are potentially endless manifestation patterns of patients with the same disorder ($2^{60} > 10^{15}$ = 10,000,000,000,000,000, assuming for simplicity binary features). Not all manifestations have different ensuing actions, but enough of them do, due to potential complications, contradicting drugs, or just 'watch out for' monitoring. Additionally, in many cases, the labels of the primary diagnoses leave some findings un-explained. Diagnosis is not complete until every abnormal finding is clinically explained, at which point we assess prognosis and decide whether or not to take treatment actions, or/and additional tests are needed to rule out a severe problem which, currently, is still a remote possibility.

Note that, in this respect, AI for medicine is unique. In many AI applications, e.g. traffic signs reading, or hand-writing digit recognition, once labeling the object is reached, the potential statistical variations in the object image, e.g. due to dirt, light conditions or hand writing, do not have impact on the implied decision of the class membership: a 'STOP' label for a traffic sign is an instruction to halt, or '6' label is a numeric 6 digit in a check amount. This article discusses AI solutions to cover the full workflow of medical diagnosis, as opposed to individual point solutions that only address narrow limited task such as MRI interpretation, or interpretation of EKG, EEG signals.

**Labeling a diagnosis is a critical step and its importance should not be under-estimated.** However, the ultimate goal of medical diagnosis is to be able to decide on the full set of the right ensuing actions to improve patient outcomes. So, how do physicians do it? What would it take to develop AI solutions to go beyond the labeling phase?

## 3. Physicians Leverage Knowledge of Anatomy, Physiology, and …

Physicians leverage their knowledge of anatomy and physiology, together with cause and effect reasoning (etiology), to come up with a more refined diagnosis - beyond just statistical labeling – that better explains the manifestation pattern of any specific patient; leading to the right ensuing actions. We will call this *complete diagnosis*. Physicians, though, do not approach diagnosis as a two-phase process with labeling being phase 1. All throughout the diagnosis process, physicians integrate statistical considerations (sensitivity/ specificity/ incidence) with anatomy and physiology considerations aiming to best understand the situation of the specific patient currently being examined.

The phrases 'every patient is unique' (or in statistical terms, a sample of one, N=1) is a viewpoint often represented in discussions on the concept of medical diagnosis. Perhaps the most salient observation supporting this notion is the place for Physician Notes in patient records. Patient records are written by physicians for physicians, implying that once you record the patient findings and the 1-5 words diagnostic label, there is no need for verbose text connecting the observed findings with the textbook description of the label. You are expected, though, to provide any additional useful information beyond the label that are unique to this specific patient and are relevant to future care.

## 4. Complete Diagnosis is Results-Oriented Diagnosis

To summarize, while labeling the diagnoses is a very important step, we cannot stop there, and we need to continue analyzing the patient situation until we reach a complete diagnosis. The complete diagnosis phase, beyond the labeling phase, is only partially a matter of extra verification, i.e. increasing the probability of the labels, notwithstanding the cost of excessive verification with minimal additional value. It is more the need for a broader/deeper patient assessment that produces specific refinements for the ensuing actions. **Medical diagnosis is only partially about probability calculations for label X or Y. It is about *clinical* understanding the overall situation of the patient for the purpose of delivering the right treatment**.

In every profession, you are ultimately measured by the **results** you deliver; not by your activities along the way. Medicine is ultimately measured by the outcomes in patient care as a result of treatment actions. Adopting the outcome-orientation approach, the diagnosis phase is completed only when it enables the best decisions that the medical profession can offer for the specific patient under consideration with his unique set of body failures. Most contemporary machine Learning models in healthcare are based on patient datasets of clinical findings and aim at diagnostic classification of IDC-10 labels or predicting clinical values. Even when patient outcomes, benefits, cost, or re-admission data are available, they, typically, are **not** included in the learning algorithms for classification or prediction of diagnosis. Few works discuss machine learning models for predicting outcomes, e.g. predict hospital re-admission, or in-hospital death. A [JAMA Aug 2019](#) [5] Viewpoint article by Shah et al eloquently describes the challenges of quantitative models to measure the impact of classification machine learning models on the actions taken by the physicians to improve patient care. It also includes references on recent works.

## 5. Beyond Knowledge-Blind Algorithms

Early successes with Deep Learning led to overstating its applicability, reaching claims such as 'with sufficient amount of data, Deep Learning can solve all AI problems'. Assuming gigantic patient datasets, can a Deep Learning algorithm learn from patient data ONLY the anatomy and physiology of the human heart? That is: learn the four chamber structure, the arteries, the valves, the conduction system and the pacemaker, the walls, …, the function of each module and the overall blood flow. I doubt it, simply because **patient data does not contain the information to enable such learning**. (To appreciate the complexity, check here for an excellent heart simulation). Architecturally, the input into case-based learning are flat vectors. Representing knowledge about body organs requires far more complex structures the like of engineering design diagrams or state machines (state charts) for real-time systems. On top of these knowledge structures we will need to add evidential and cause-and-effect reasoning; two areas where current data-centric machine learning algorithms have inherent limitations, as they are only based on reasoning by similarity and extrapolation, Pearl [6].

By analogy, with data about billions of "things", and thousands of apples that fall down every day, deep learning algorithms, with no human touch, can certainly come up with a model to calculate the time at which any given falling object will touch the ground. But can data-only deep learning model come up with Newton's laws? I mean produce models that represent deeper generalized **understanding of the Universe forces** along with "compact" formula such as Time to hit the ground=SQRT(2*Height/g), where g=gravity=9.8m/s$^2$ (as opposed to a gigantic black box neural net)?

**Machines can learn many things from data, but data is not the only source that machines can learn from. Historic patient data only tells us *what* the possible manifestations of a certain body failure are. Anatomy and physiology knowledge tell us *how* the body works and fails. Both are needed for complete diagnosis.**

With the understanding that **disorder labeling via ML and medical diagnosis are not quite the same**, I propose the Double Deep Learning [7] approach which advocates integrating **machine-teaching** of **deep knowledge** with contemporary **data-only machine-self-learning** techniques. The term 'deep knowledge' refers to the difference between teaching physicians versus teaching paramedics, where, with physicians, we teach deep fundamental principles, and models, like anatomy and physiology and the necessary parts of chemistry and biology. This enables deeper reasoning beyond reasoning by similarity and extrapolation of contemporary data-centric knowledge-blind machine learning algorithms. Machine-teaching of **deep knowledge** is also in contrast to teaching shallow prescriptive knowledge in the form of explicit IF-THEN rules connecting facts to conclusions that characterized the classic rule-based systems of the early AI days. Way back (1978), I discussed the limitations of rule-based systems specifically in the context of medical diagnosis applications [8].

Additionally, by their very nature, statistics-based machine learning algorithms entail error probabilities which, for complex cases, could be quite high. Reducing this error probability by adding to AI systems reasoning which is based on **complimentary knowledge sources** could benefit greatly the diagnostic accuracy of the labeling phase and, as explained above, the depth and breadth of the complete diagnosis.

## 6. Wikipedia for Smart Machines

In [7] I propose an initiative to build Wikipedia for Smart Machines, meaning targeting readers that are **not** human, but rather smart machines. Named **ReKopedia**, e.g. **Medical ReKopedia,** the goal is to develop methodologies, tools, and algorithms to convert medical knowledge that we learn in schools, universities and during our professional life into **Re**usable **Kno**wledge structures that *smart machines* can use in their inference algorithms. Ideally, ReKopedia would be an open source shared knowledge repository similar to the well-known shared open source software code repositories. Reviewing the **syllabuses of schools and universities** in different areas will teach us the content we teach humans and can be a good starting point for the content we should teach machines. For medicine, the Physiome Project is a great initiative to develop standards, models and databases for computational modelling of human organs. See for instance, the 2016 special edition of The Journal of Physiology [9] which focuses on the Cardiac Physiome Project and provides additional references. The results coming out from this important project, with human as target readers, could be a great source for structuring medical knowledge into **Medical ReKopedia** modules to be used by AI algorithms in medical support systems. Investment in the ReKopedia initiative has tremendous value for society, because once knowledge packages for smart machines exist, an unlimited number of computers that use it can be operational in seconds. To produce an expert physician requires nine months followed by at least 30 years, and then all is gone after a limited, and unpredictable, number of years.

## 7. Summary

To summarize, data-centric machine learning algorithms have a convincing proven record, will continue to evolve, and have their place in the architecture of every AI solution for medical decision support. Nonetheless, like any other mathematical technique, they have their limitations and applicability scope. The goal of this article is to complement them with techniques to overcome their **inherent** limitations as knowledge-blind algorithms. **The Double Deep Learning approach, along with the initiative for Wikipedia for smart machines, leads to AI diagnostic support solutions for Complete Diagnosis beyond the limited data-only labeling solutions we see today.** The clinical value and business value of AI systems for medicine will forever be limited until they also 'understand' the content we teach medical students in universities.

## About the Author

*Professor Moshe BenBassat has been researching, practicing and teaching Artificial Intelligence, from the first AI "Spring" of the 1980's, all through the "Winter" of the 1990s and early decade of this century, and now into the AI renaissance that is blossoming again. He is one the pioneers in developing AI-based support software for medical diagnosis and treatment; including for Endocrinology/Infertility, Emergency and Critical Care (MEDAS), Space Medicine, and Arthritis. By 1990, long after he moved on, MEDAS reached a "90% level agreement with the gold standard diagnosis of the attending physician" and grew to cover 350 disorders by means of 6000 features organized in hierarchical Bayesian networks.*

*During a long academic career with positions at Tel Aviv University, USC, UCLA, and now at IDC Herzliya, Professor BenBassat made many contributions in Pattern Recognition, Artificial Intelligence, Optimization, Data Science and Machine Learning.*

*Following his invention of "service chain optimization" (patent awarded), he founded ClickSoftware which has been leading this space since its inception in 1997. Moshe served as ClickSoftware's CEO until 2015, at which point it was acquired by a private equity firm. Moshe also founded Plataine which is focused on intelligent automation for smart manufacturing, leveraging Industrial Internet of Things (IIoT) and Artificial Intelligence technologies.*